\documentclass{elsarticle}

\usepackage{graphics}
\usepackage{graphicx}
\usepackage{epsfig}
\usepackage{epstopdf}

\usepackage{amsmath}
\usepackage{amssymb}
\usepackage{latexsym}
\newtheorem{thm}{Theorem}

\newtheorem{defn}[thm]{Definition}

\usepackage{algorithm}
\usepackage{algorithmic}
\usepackage{subfig}
\usepackage{colortbl}
\usepackage{multirow}

\begin{document}

\begin{frontmatter}



\title{
Cost-sensitive C4.5 with post-pruning and competition
}
\author{Zilong Xu}
\ead{xzl-wy163@163.com}
\author{Fan Min\corref{cor1}}
\ead{minfanphd@163.com}
\author{William Zhu}
\ead{williamfengzhu@gmail.com}

\cortext[cor1]{Corresponding author. Tel.: +86 133 7690 8359}

\address{Lab of Granular Computing,
Zhangzhou Normal University, Zhangzhou 363000, China}


\begin{abstract}
Decision tree is an effective classification approach in data mining and machine learning.
In applications, test costs and misclassification costs should be considered while inducing decision trees.
Recently, some cost-sensitive learning algorithms based on ID3 such as CS-ID3, IDX, $\lambda$-ID3 have been proposed to deal with the issue.
These algorithms deal with only symbolic data.
In this paper, we develop a decision tree algorithm inspired by C4.5 for numeric data.
There are two major issues for our algorithm.
First, we develop the test cost weighted information gain ratio as the heuristic information.
According to this heuristic information, our algorithm is to pick the attribute that provides more gain ratio and costs less for each selection.
Second, we design a post-pruning strategy through considering the tradeoff between test costs and misclassification costs of the generated decision tree.
In this way, the total cost is reduced.
Experimental results indicate that (1) our algorithm is stable and effective;
(2) the post-pruning technique reduces the total cost significantly;
(3) the competition strategy is effective to obtain a cost-sensitive decision tree with low cost.
\end{abstract}

\end{frontmatter}

  %
  %
\section{Introduction}\label{section: introduction}

In data mining and machine learning, there are many methods such as artificial neural networks \cite{mcculloch1943logical,ZhouZ2006}, Bayesian networks \cite{ChaiX2004}, Rough sets \cite{Pawlak02Rough}, etc.
Decision tree induction \cite{quinlan1986induction,Quinlan93C45} is one of the simplest, and yet most successful one \cite{RussellS2009Artificial}.
Test costs and misclassification costs are two important types of costs in applications.
We should consider both of them while inducing decision trees.
Recently, some cost sensitive decision tree algorithms based on ID3 such as IDX \cite{NortonSW1989Generating}, CS-ID3 \cite{Tan1993Cost}, $\lambda$-ID3 \cite{Min2012Competition} have proposed to deal with the issue.
However, the existing ones deal with only symbolic data, while there are a number of numeric data in applications.

In this paper, we develop an algorithm named Competition Cost-sensitive C4.5 for numeric data.
This algorithm is inspired by C4.5 \cite{Quinlan93C45}.
C4.5 is based on the information gain ratio.
In contrast, our algorithm uses the test cost weighted information gain ratio as the heuristic information.
We adopt a parameter $\lambda$ as the exponent of the test cost.
In the idea is to pick the attribute which provides more information gain ratio and costs less for each selection
When we implement the algorithm, we let the test cost of the tested attribute be one unit to avoid considering the test cost in the heuristic information repeatedly.

We develop a post-pruning technique considering the tradeoff between test costs and misclassification costs.
After a decision tree is built, our algorithm traverses it in post order to prune.
First, compute the average cost of the subtree corresponding to current non-leaf node.
Second, try to cancel the attribute test represented by the current node.
Third, compute the average cost of the subtree after post-pruning.
If this average cost is less than the former, we cut current node, otherwise reserve the current node.
Through post-pruning, the average cost of obtained decision trees are reduced.

We represent the specific procedures of our algorithm as follows.
Step 1, produce a population of decision trees according to different parameter settings.
Step 2, post prune these decision trees.
Step 3, compute average cost including test cost and misclassification cost of each decision tree in the population.
Step 4, select the tree with minimal average cost as the result to output.

Our algorithm is tested on University of California - Irvine (UCI) datasets.
Various parameter settings on both test costs and misclassification costs are used to simulate different situation of applications.
We implement the algorithm with Java in our open source software Cost-sensitive Rough Set (COSER) \cite{Coser}.
Three representative distributions, namely Uniform, Normal and Pareto, are employed to generate test costs.
Experimental results indicate that: (1) our algorithm is stable and effective without overfitting; (2) no $\lambda$ always beats others on different test cost settings, hence it is not good to use a fixed $\lambda$ setting; (3) the decision tree with low cost on training dataset tends to have low cost on testing dataset; (4) post-pruning strategy helps reduce the average cost significantly.

The rest of this paper is organized as follows.
Section \ref{section: Preliminaries} reviews the basic knowledge involved in our paper.
Section \ref{section:algorithm} proposes our decision tree algorithm to tackle numeric data issue.
In section \ref{section:example}, we explain the process of our algorithm through a simple example.
We reveal experimental results and analysis in Section \ref{section:Experiment}.
The conclusion is represented in Section \ref{section: conclusion}.

  %
  %
\section{Preliminaries}\label{section: Preliminaries}

This section reviews the basic knowledge including C4.5 algorithm, cost sensitive decision system, and computation of average cost involved in this paper.

\subsection{C4.5 algorithm}

C4.5 is a suite of algorithms for classification problems in machine learning and data mining \cite{Wu2009Top}.

It is targeted at supervised learning: Given an attribute-valued dataset where instances are described by collections of attributes and belong to one of a set of mutually exclusive classes, C4.5 learns a mapping from attribute values to classes that can be applied to classify new, unseen instances \cite{Wu2009Top}.

The heuristic information is
\begin{equation}\label{equation:gainratio}
    GainRatio(a)=\frac{Gain(a)}{Split\_infor(a)}.
\end{equation}

\subsection{The decision system with test costs and misclassification costs}\label{subsection: bcds}

We consider decision systems with test costs and misclassification costs.
Since the paper is the first step toward this issue, we consider only simple situations.

\begin{defn} \cite{MinZhu11MinimalCost}
A decision system with test costs and misclassification costs (DS-TM) S is the 7-tuple:
\begin{equation}\label{equation:bdcds}
S=(U,C,d,V,I,tc,mc),
\end{equation}
where U is a finite set of objects called the universe, C is the set of conditional attributes, d is the decision attribute, $V = \{ {V_a}|a \in C \cup \{ d\} \} $ where $V_a$ is the set of values for each $a \in C \cup \{ d \}$, $I = \{ I_a | a \in C \cup \{d\}\}$ where $I_a:U \rightarrow V_a$ is an information function for each $a \in C \cup \{ d\}, tc:C\rightarrow \mathbb{R}^{+} \cup \{ 0\}$ is the test cost function, and $mc: k \times k \rightarrow \mathbb{R}^{+} \cup \{0\}$ is the misclassification cost function, where $\mathbb{R}^{+}$ is the set of positive real numbers, and $k=|I_d|$.
\end{defn}

\subsection{The average cost}
We define the average cost of a decision tree as follows.
Let $T$ be a decision tree, $U$ be the testing dataset, and $x \in U$.
$x$ follows a path from the root of $T$ to a leaf.
Let the set of attributes on the path be $A(x)$.
The test cost of $x$ is $tc(A(x))$.
Let the real class label of $x$ be $C(x)$ and $x$ is classified as $T(x)$.
The misclassification cost of $x$ is $mc(C(x), T(x))$.
Therefore the total cost of $x$ is $tc(A(x))+mc(C(x), T(x))$.
\begin{defn}\cite{Min2012Competition}
The average cost of $T$ on $U$ is
\begin{equation}\label{equation:averagecost}
{T_C}(U) = \sum\limits_{x \in U} {(tc(A(x)) + mc(C(x)),T(x)))/|U|}.
\end{equation}
\end{defn}

  %
  %
\section{The algorithm}\label{section:algorithm}

We develop a decision tree algorithm based on C4.5.
Our algorithm is illustrated as follows.

\subsection{The heuristic information}

We present the following heuristic function:

\begin{equation}\label{equation: heuristic information}
    f(a)=GainRatio(a)[tc(a)]^{\lambda}~~~~(\lambda \leqslant 0),
\end{equation}
where $a$ is an attribute of the given decision system, $GainRatio(a)$ is the information gain of the attribute $a$ with the same meaning in C4.5 algorithm, $\lambda$ is the exponent value of the test cost of the attribute $a$.

According to the heuristic function, our algorithm selects the attribute which provides more classification accuracy.
At the same time, attributes with high test cost will be punished.

A numeric attribute can be used to classify once more in C4.5.
In practice, after measuring an attribute, we have known and recorded its value, and does not test it again.
So considering test cost of the used attribute in heuristic information is not reasonable.
When the attribute $a$ is used again, we let $tc(a)=1$ in Equation (\ref{equation: heuristic information}) to avoid computing the test cost repeatedly.
If we let $tc(a)=0$, the heuristic function has no mathematical meaning, then we use one unit to replace zero.
This operation means that our new heuristic information degenerates into the traditional heuristic information shown in Equation (\ref{equation:gainratio}) in this case.

  %
  %
\subsection{Post-pruning technique}
We employ the following post-pruning technique.
After the cost-sensitive decision tree is built, our algorithm traverses and prunes it in post-order.
The essence of the pruning technique is to judge if a test can be canceled.
Its purpose is to reduce the total cost of cost-sensitive decision tree.
When the average cost of a subtree after canceling the test is less than the one reserving the test, the attribute test will be canceled.

We illustrate the post-pruning operations through a running example in Section \ref{section:example}.
For evaluating the effect of the pruning technique, we propose measures as follows.

\begin{defn}
The reduction ratio of the cost is
\begin{equation}\label{equation:ReduceRatio}
    r=\frac{ac-ac'}{ac},
\end{equation}
where $ac$ is the average cost of the initial tree before pruning, $ac'$ is the average cost of the pruned tree.
\end{defn}

\begin{defn}
The average reduction ratio of the average cost is
\begin{equation}\label{equation:AverageReduceRatio}
ar = \frac{1}{n}\sum\limits_{i = 1}^n {{r_i}},
\end{equation}
where $n$ is the number of the parameter $\lambda$, each $r_i$ is the reduce ratio of average cost of decision trees corresponding to each $\lambda$.
\end{defn}

For example, the average cost of the decision tree before pruning is 100, and the average cost after pruning is 60, then the reduction ratio of the cost is $r=(100-60)/100=40\%$.
Suppose the reduction of cost corresponding to each $\lambda$ setting are 40\%, 50\%, 30\%, and 20\%, the average reduction ratio of the cost is $ar=(40\%+50\%+30\%+20\%)/4=35\%$.

\subsection{The competition approach}

Our algorithm adopts a competition approach to select the decision tree with the best performance on the training dataset.
It works as follows:

\begin{itemize}
  \item Step 1. Generate a batch of decision trees according to different lambda values.
  \item Step 2. Compute the average cost of each obtained decision tree.
  \item Step 3. Select the tree with minimal average cost among the batch for classification.
\end{itemize}

  %
  %
\section{A running example}
\label{section:example}

We illustrate our algorithm through a simple example.
A simple numeric decision system is given by Table \ref{table:example}.
This decision system is a sample of the dataset \texttt{diabetes}.
There are 8 condition attributes and 24 instances in the decision system.
All the condition attributes are continuous value.
The $\lambda$ value is set to be -2 in this example.

\begin{table}[tb]\caption{An example of numeric decision system}
\label{table:example}
\centering
\setlength{\tabcolsep}{12pt}
\begin{tabular}{ccccccccc}
  \hline
  $a_1$ & $a_2$ & $a_3$ & $a_4$ & $a_5$ & $a_6$ & $a_7$ & $a_8$ & D \\
  \hline
2 & 100 & 68 & 25 & 71 & 38.5 & 0.324 & 26 & 0\\
15 & 136 & 70 & 32 & 110 & 37.1 & 0.153 & 43 & 1\\
1 & 107 & 68 & 19 & 0 & 26.5 & 0.165 & 24 & 0\\
1 & 80 & 55 & 0 & 0 & 19.1 & 0.258 & 21 & 0\\
4 & 123 & 80 & 15 & 176 & 32 & 0.443 & 34 & 0\\
7 & 81 & 78 & 40 & 48 & 46.7 & 0.261 & 42 & 0\\
4 & 134 & 72 & 0 & 0 & 23.8 & 0.277 & 60 & 1\\
2 & 142 & 82 & 18 & 64 & 24.7 & 0.761 & 21 & 0\\
6 & 144 & 72 & 27 & 228 & 33.9 & 0.255 & 40 & 0\\
2 & 92 & 62 & 28 & 0 & 31.6 & 0.13 & 24 & 0\\
1 & 71 & 48 & 18 & 76 & 20.4 & 0.323 & 22 & 0\\
6 & 93 & 50 & 30 & 64 & 28.7 & 0.356 & 23 & 0\\
1 & 122 & 90 & 51 & 220 & 49.7 & 0.325 & 31 & 1\\
1 & 163 & 72 & 0 & 0 & 39 & 1.222 & 33 & 1\\
1 & 151 & 60 & 0 & 0 & 26.1 & 0.179 & 22 & 0\\
0 & 125 & 96 & 0 & 0 & 22.5 & 0.262 & 21 & 0\\
1 & 81 & 72 & 18 & 40 & 26.6 & 0.283 & 24 & 0\\
2 & 85 & 65 & 0 & 0 & 39.6 & 0.93 & 27 & 0\\
1 & 126 & 56 & 29 & 152 & 28.7 & 0.801 & 21 & 0\\
1 & 96 & 122 & 0 & 0 & 22.4 & 0.207 & 27 & 0\\
1 & 128 & 98 & 41 & 58 & 32 & 1.321 & 33 & 1\\
8 & 109 & 76 & 39 & 114 & 27.9 & 0.64 & 31 & 1\\
5 & 139 & 80 & 35 & 160 & 31.6 & 0.361 & 25 & 1\\
3 & 111 & 62 & 0 & 0 & 22.6 & 0.142 & 21 & 0\\
  \hline
\end{tabular}
\end{table}

Suppose the test cost of each attribute is a random integer obeying uniform distribution in [1, 10].
The test cost vector is listed as Table \ref{table:TestCostSetting}.

\begin{table}[tb]\caption{A test cost vector}
\label{table:TestCostSetting}
\setlength{\tabcolsep}{12pt}
\begin{center}
\begin{tabular}{ccccccccc}
\hline
attribute   & $a_1$ & $a_2$ & $a_3$ & $a_4$ & $a_5$ & $a_6$ & $a_7$ & $a_8$\\
\hline
$tc$        & 4 & 1 & 4 & 1 & 7 & 7 & 8 & 9\\
\hline
\end{tabular}
\end{center}
\end{table}

The misclassification cost matrix is set to be
$\left[ {\begin{array}{*{20}{c}}
   0 & {500}  \\
   {50} & 0  \\
\end{array}} \right]$
.

\textbf{Step 1 Build the tree through selecting attribute}

\begin{figure}[tb]
\centering
\subfloat[The initial tree]{
    \label{fig:prune01}
    \includegraphics[scale=0.4]{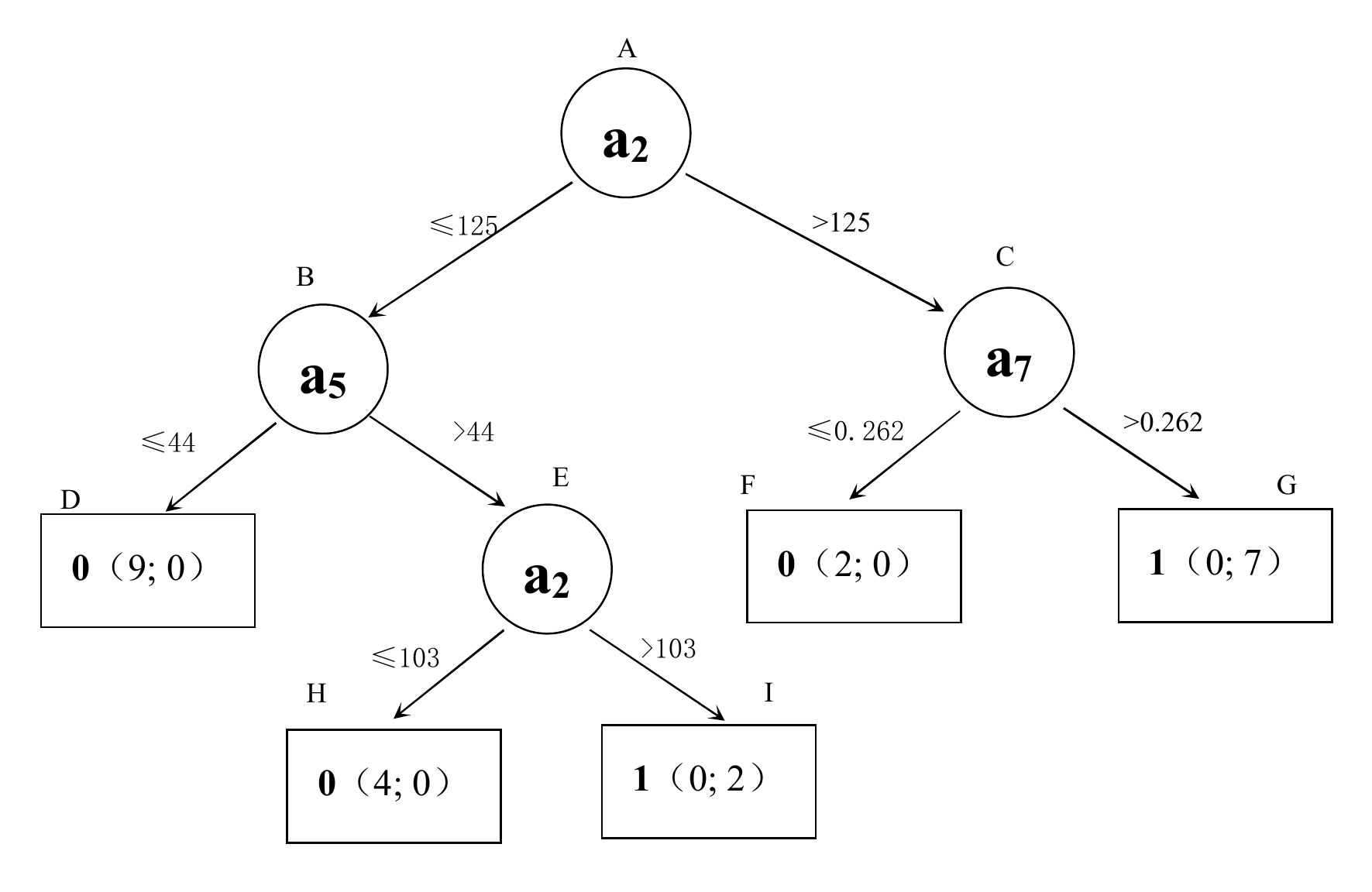}
}
\subfloat[Cut the subtree E]{
    \label{fig:prune02}
    \includegraphics[scale=0.4]{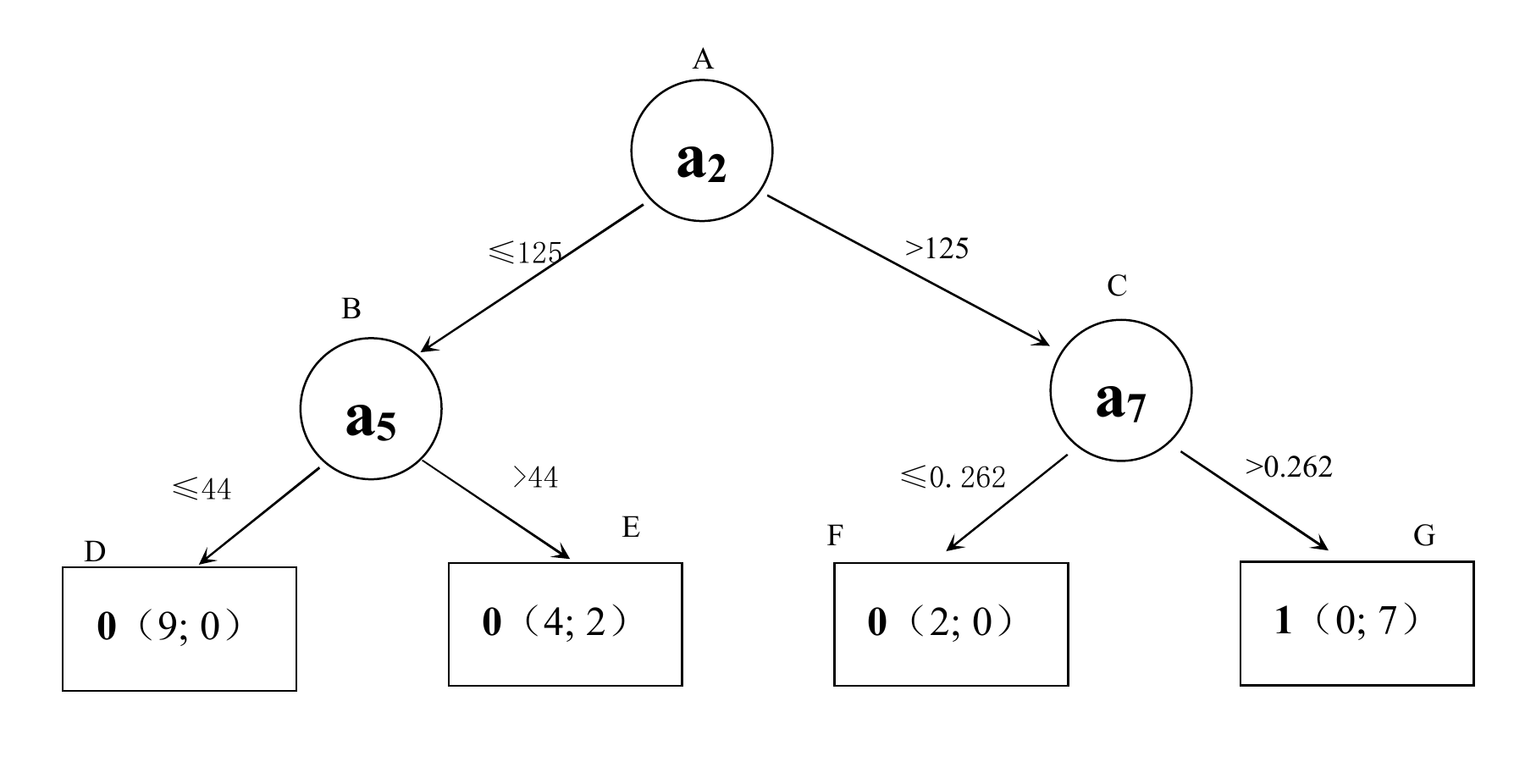}
}\\
\subfloat[Cut the subtree B]{
    \label{fig:prune03}
    \includegraphics[scale=0.4]{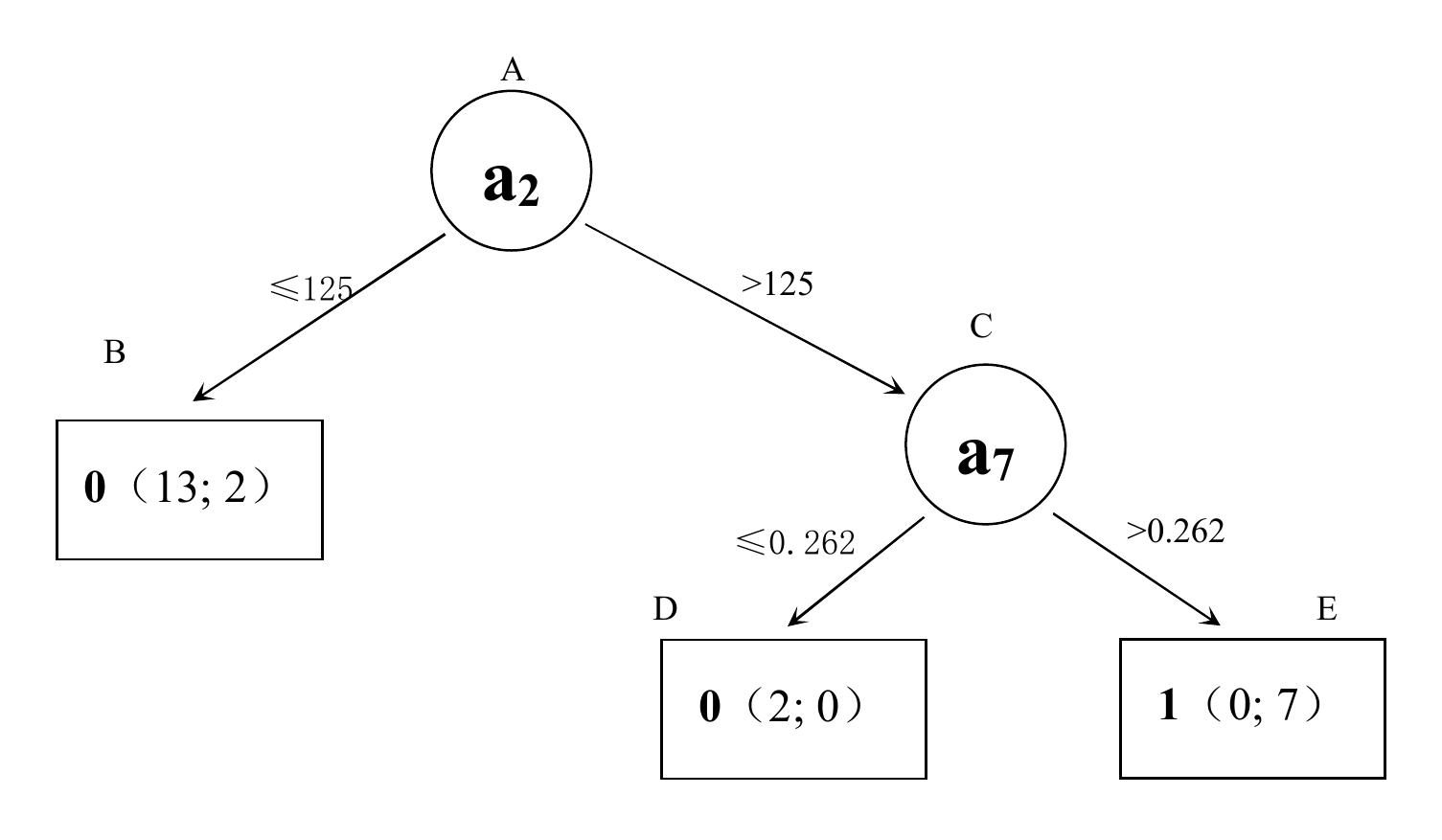}
}
\subfloat[Cut the subtree C]{
    \label{fig:prune04}
    \includegraphics[scale=0.4]{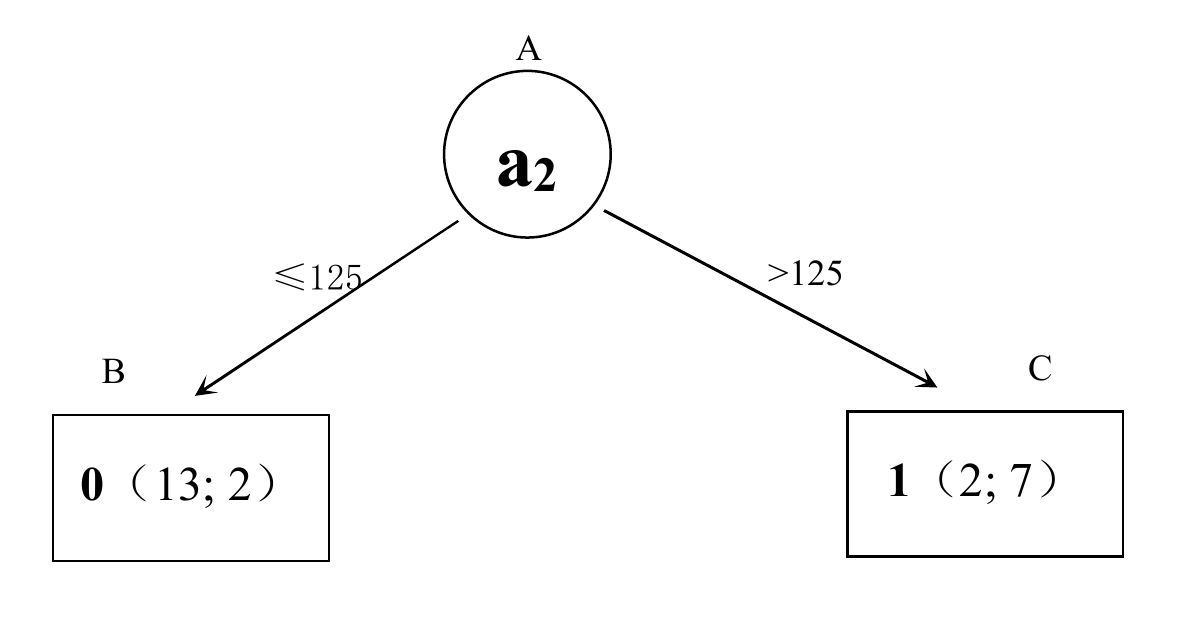}
}\\
\subfloat[Cut the subtree A]{
    \label{fig:prune05}
    \includegraphics[scale=0.4]{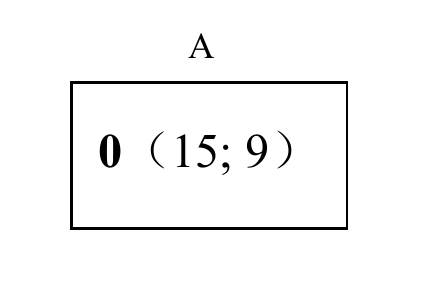}
}
\caption{Post-pruning process}
\label{fig:Post-pruning}
\end{figure}

Since C4.5 is not the keynote in this paper, we omit some details involving original C4.5.
The difference between our algorithm and C4.5 is the heuristic information.
The initial decision tree is shown as Figure \ref{fig:prune01}.

\textbf{Step 2 Post-pruning}

After obtained the initial decision tree, the algorithm will post prune it to reduce its average cost.

The initial decision tree is revealed in Figure \ref{fig:prune01}.
Because each instance is classified correctly, there is no misclassification cost, $mc=0$.
To classify all instances in the node D, the attribute $a_2$ and $a_5$ must be tested.
Then the total test cost of node is D is $tc_B=(1+7) \times 9=72$.
Similarly, the total test cost of the node H, I,F and G are $(1+7) \times 4=32$, $(1+7) \times 2=16$, $(1+8) \times 2=18$ and $(1+8) \times 7=63$ respectively.
Then the average cost of the tree is $(tc_D+tc_H+tc_I+tc_F+tc_G)/|U|=(72+32+16+18+63)/24=8.375$

Post-pruning runs in post order.
We represent the specific procedures by following text and Figure \ref{fig:Post-pruning}.

(1) The test represented by the node E is firstly considered.
We try to cut the subtree whose root node is E representing attribute $a_2$.
The tree after cutting is shown as Figure \ref{fig:prune02}.
Let we compute the average cost of the tree illustrated by Figure \ref{fig:prune02}.
In the leaf node E, there are 4 instances belonging to class 0 and 2 instances belonging to class 1.
The predict class is 0, because instances whose class is 0 are more than instances whose class is 1.
Then, we can say that there are 4 instances are classified correctly and other 2 instances are misclassified.
So, the total misclassification cost of the leaf node E is $mc_E=50 \times 2=100$.
The total test cost of the node E is $tc_E=(1+7) \times 6=48$.
Then, the average cost of the subtree whose root is the node E in Figure \ref{fig:prune02} is $ac_E=(tc_E+mc_E)/|U_E|=(48+100)/6=24.67$.
Before pruning the subtree E, the average cost of the subtree E is $(32+16)/6=8$ shown in \ref{fig:prune01}.
Obviously, this cost is more than the cost of the initial tree, so the test of the node E can not be canceled.

(2) Then we consider the attribute test $a_5$ represented by the node B.
The average cost of the initial subtree whose root is node B means $(tc_D+tc_H+tc_I)/|U_B|=(72+32+16)/15=8$ in Figure \ref{fig:prune01}.
We try to cut the subtree whose root is the node B.
The tree after cutting is shown as Figure \ref{fig:prune03}.
In the leaf node B, there are 13 instances belonging to class 0 and other 2 instances belonging to class 1.
The predict class is 0, because instances whose class is 0 are more than instances whose class is 1.
Then, we can say that there are 13 instances are classified correctly and other 2 instances are misclassified.
So, the total misclassification cost of the leaf node C is $mc_C=50 \times 2=100$.
The total test cost of the leaf node B is $1 \times 15=15$.
Then, the the average cost of the subtree whose root is the node C in Figure \ref{fig:prune03} is $(tc_B+mc_B)/|U_B|=(15+100)/15=7.67$.
The algorithm cancel the test of attribute $a_5$ represented by node B.
The pruned tree is revealed by Figure \ref{fig:prune03}.

(3) We consider the attribute test represented by the node A.
The average cost of the initial subtree whose root is node C means $(tc_F+tc_G)/|U_C|=(18+63)/9=9$ in Figure \ref{fig:prune01}.
We try to cut the subtree whose root is the node C.
The tree after cutting is shown as Figure \ref{fig:prune04}.
In the leaf node C, there are 2 instances belonging to class 0 and other 7 instances belonging to class 1.
The predict class is 1, because instances whose class is 1 are more than instances whose class is 0.
Then, we can say that 7 instances are classified correctly and other 2 instances are misclassified.
So, the total misclassification cost of the leaf node A is $mc_C=500 \times 2=1000$.
Then, the average cost of the subtree is $(tc_C+mc_C)/|U|=(9+1000)/9=112.1$.
Obviously, this cost is more than that of the initial subtree, so we can not cancel the attribute test $a_7$ represented by the node C.

(4) Finally, we consider the attribute test $a_2$ represented by the node A.
We try to cut the subtree whose root is node A.
The tree after cutting is shown as Figure \ref{fig:prune05}.
In the leaf node A, there are 15 instances belonging to class 0 and other 9 instances belonging to class 1, so predict class is 0.
Then, we can say that there are 15 instances are classified correctly and other 2 instances are misclassified.
So, the total misclassification cost of the leaf node A is $mc_C=50 \times 9=450$.
The tree is only one leaf node, thus no test is taken, the total test cost is 0.
Then, the average cost of the subtree is $mc_a/|U|=450/24=18.75$.
We can conclude the attribute test $a_2$ can not be canceled.

After post-pruning, the final decision tree is illustrated in Figure \ref{fig:prune01}.
We also reveal the whole process of post-pruning in the Table \ref{table:PruneProcess}.

\textbf{Step 3 Competition}
In this example, the $\lambda$ is set to be -2.
If we use the competition approach, the algorithm produces a batch of decision trees by different $\lambda$ values on training dataset.
These trees can be post-pruning or not.
Compute the average cost of each tree and select the one with minimal cost among these trees as the competition cost-sensitive decision tree (CC-Tree).

\begin{table}\caption{The post-pruning process of the running example}
\centering
\label{table:PruneProcess}
\begin{tabular}{cccccccccc}
  \hline
\multirow{2}{*}{Step} & \multirow{2}{*}{Attribute test node} & \multicolumn{3}{c}{cost (no pruning)} & \multicolumn{3}{c}{cost (prune)} & \multirow{2}{*}{instance count} & \multirow{2}{*}{prune?}\\
 &  & tc & mc & ac & tc & mc & ac &  &\\
  \hline
1 & E(a2) & 48 & 0 & 8 & 48 & 100 & 24.67 & 6 & No\\
2 & B(a5) & 120 & 0 & 8 & 15 & 100 & 7.667 & 15 & Yes\\
3 & C(a7) & 81 & 0 & 9 & 9 & 1000 & 112.1 & 9 & No\\
4 & A(a2) & 201 & 0 & 8.375 & 0 & 450 & 18.75 & 24 & No\\
  \hline
\end{tabular}
\end{table}

  %
  %
\section{Experiments}\label{section:Experiment}

In this section, we try to answer the following questions by experimentation.

(1) Is our algorithm stable?

(2) Is there optimal settings of parameter $\lambda$?

(3) Does the competition strategy adopted by CC-C4.5 improve the performance over any particular decision tree?

(4) Can the post-pruning strategy improve the performance of decision trees?

We use 60\% of the dataset as the training data, and then the remaining part as the testing data.
The test cost of each attribute is set to a random integer in [1, 10].


\subsection{Experiment results}

\subsubsection{Parameter comparison}

\begin{figure}[tb]
\centering
\subfloat[]{
    \label{fig:ParaComU}
    \includegraphics[scale=0.45]{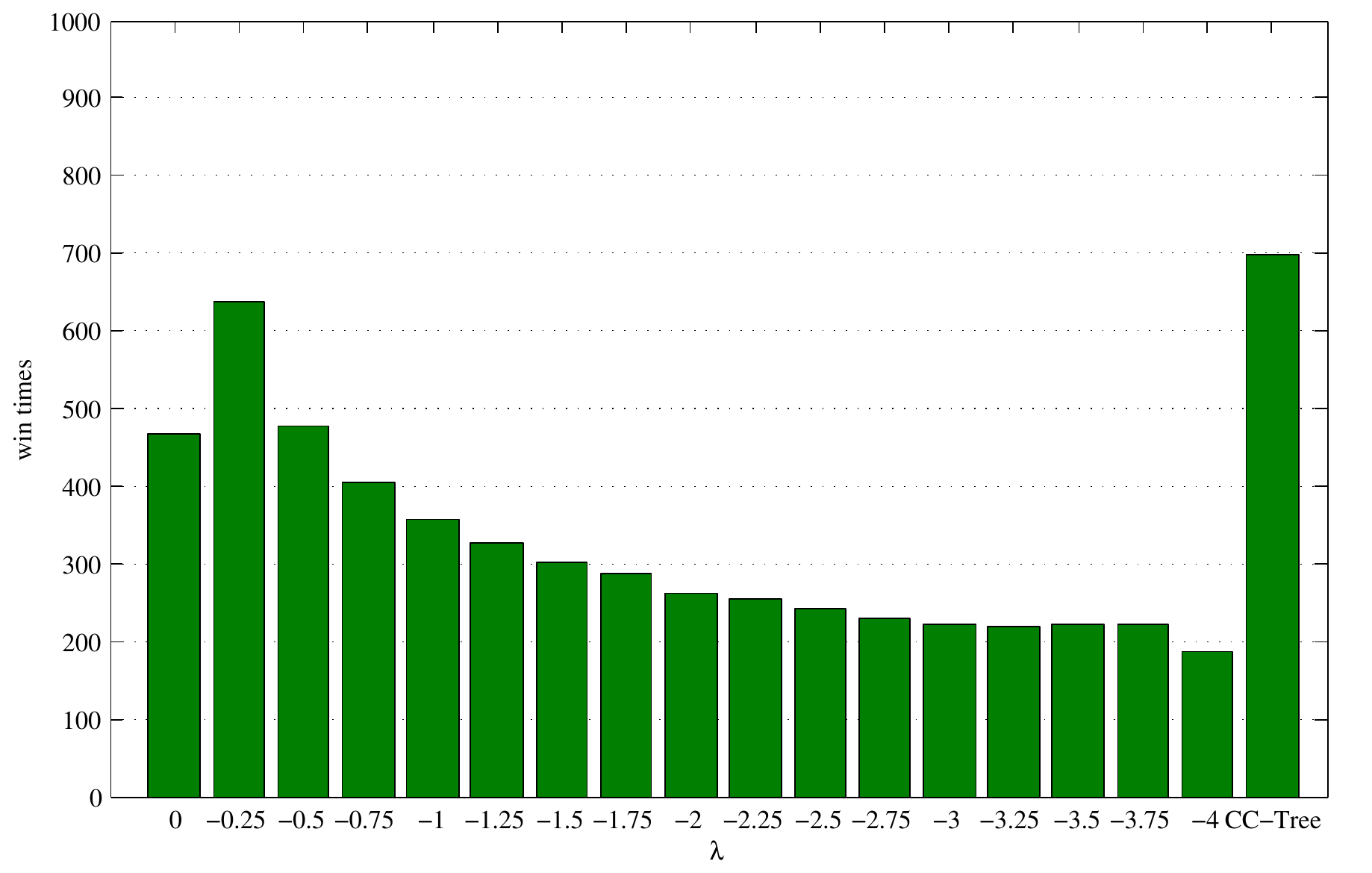}}\\
\subfloat[]{
    \label{fig:ParaComN}
    \includegraphics[scale=0.45]{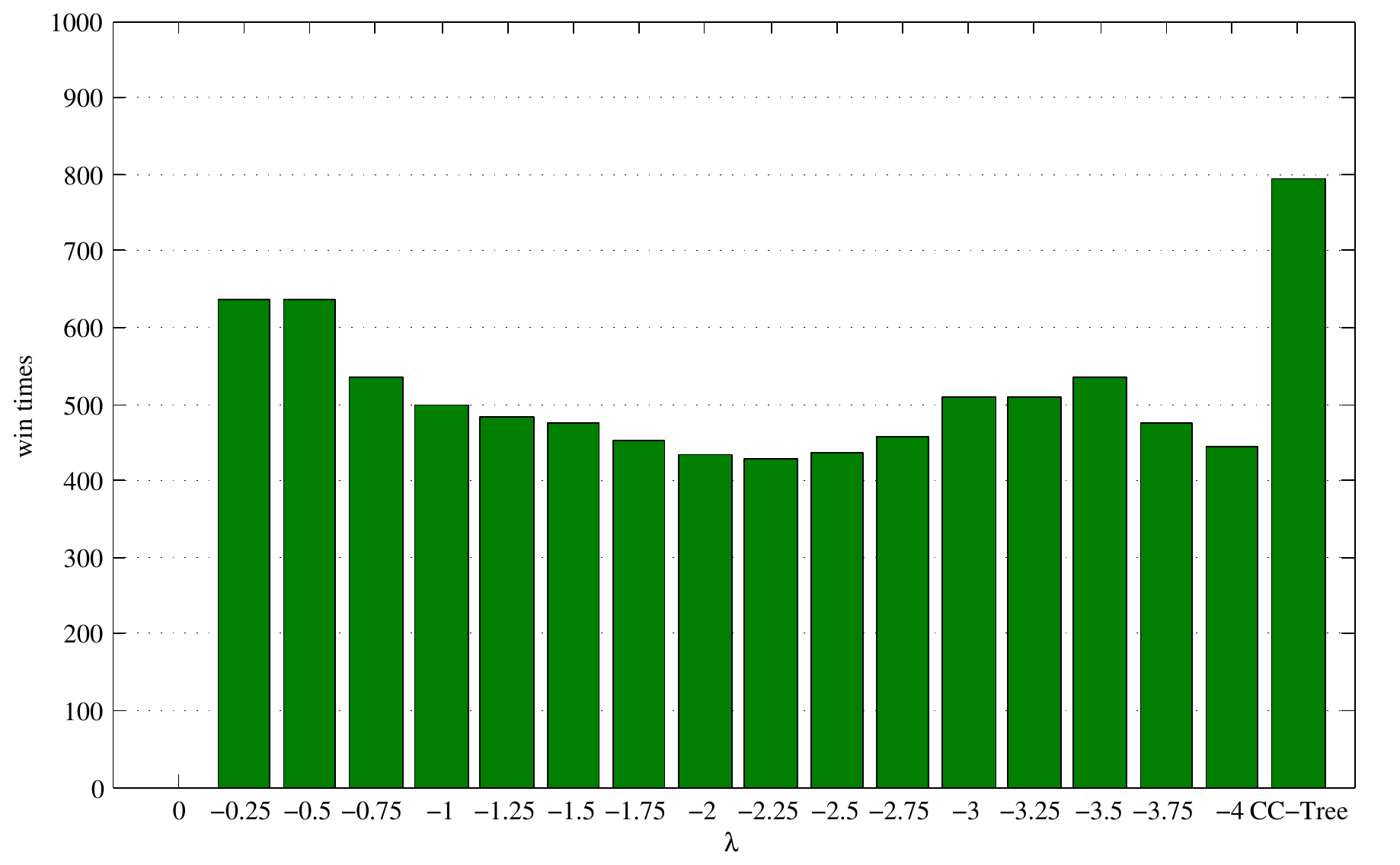}}\\
\subfloat[]{
    \label{fig:ParaComP}
    \includegraphics[scale=0.45]{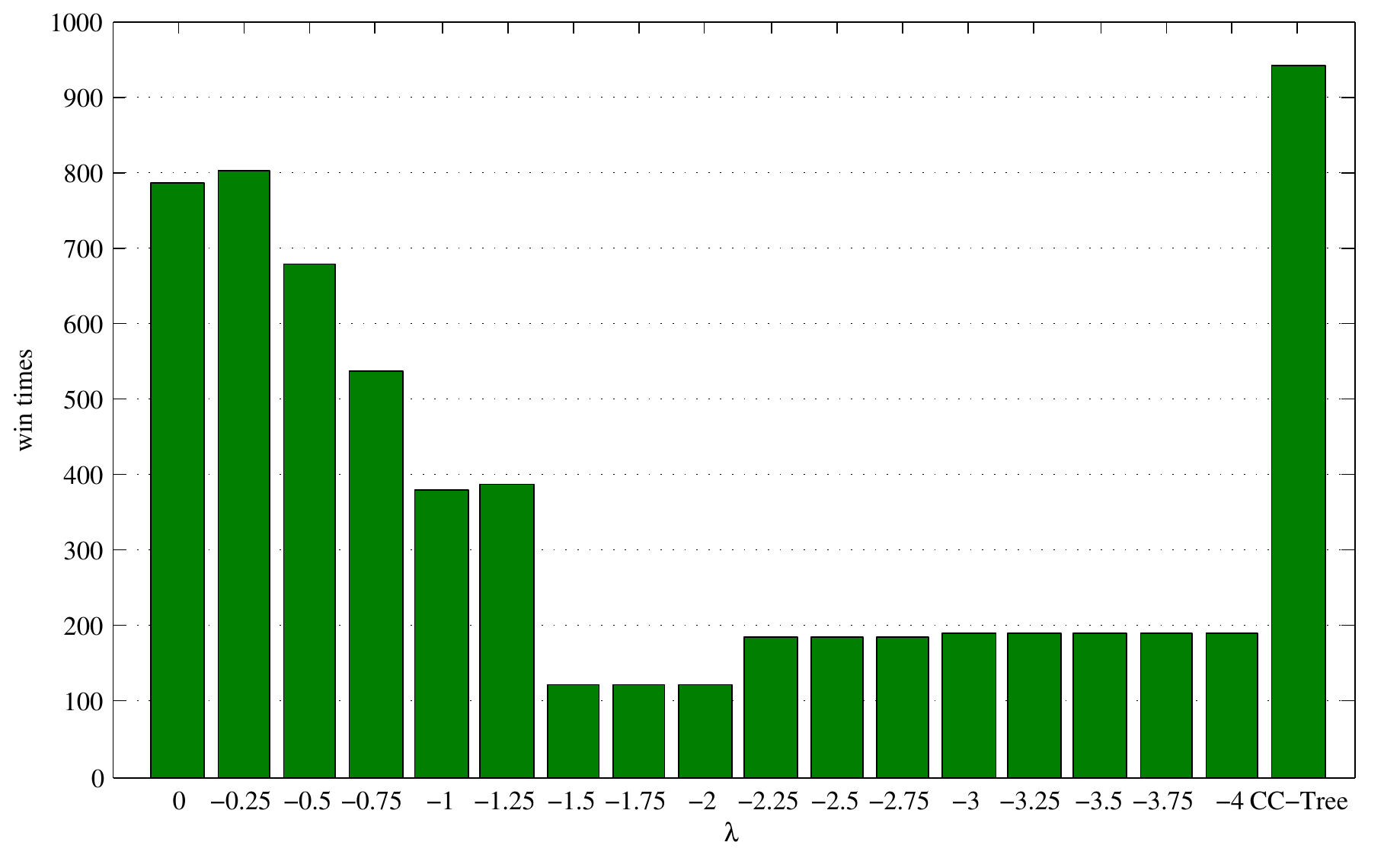}}\\
\caption{Parameter comparison on \texttt{diabetes} dataset with three distributions: (a) Uniform (b) Normal (c) Pareto} \label{fig:ParaComparWintime}
\end{figure}

\begin{figure}[tb]
\centering
\subfloat[]{
    \label{fig:ParaComCostU}
    \includegraphics[scale=0.45]{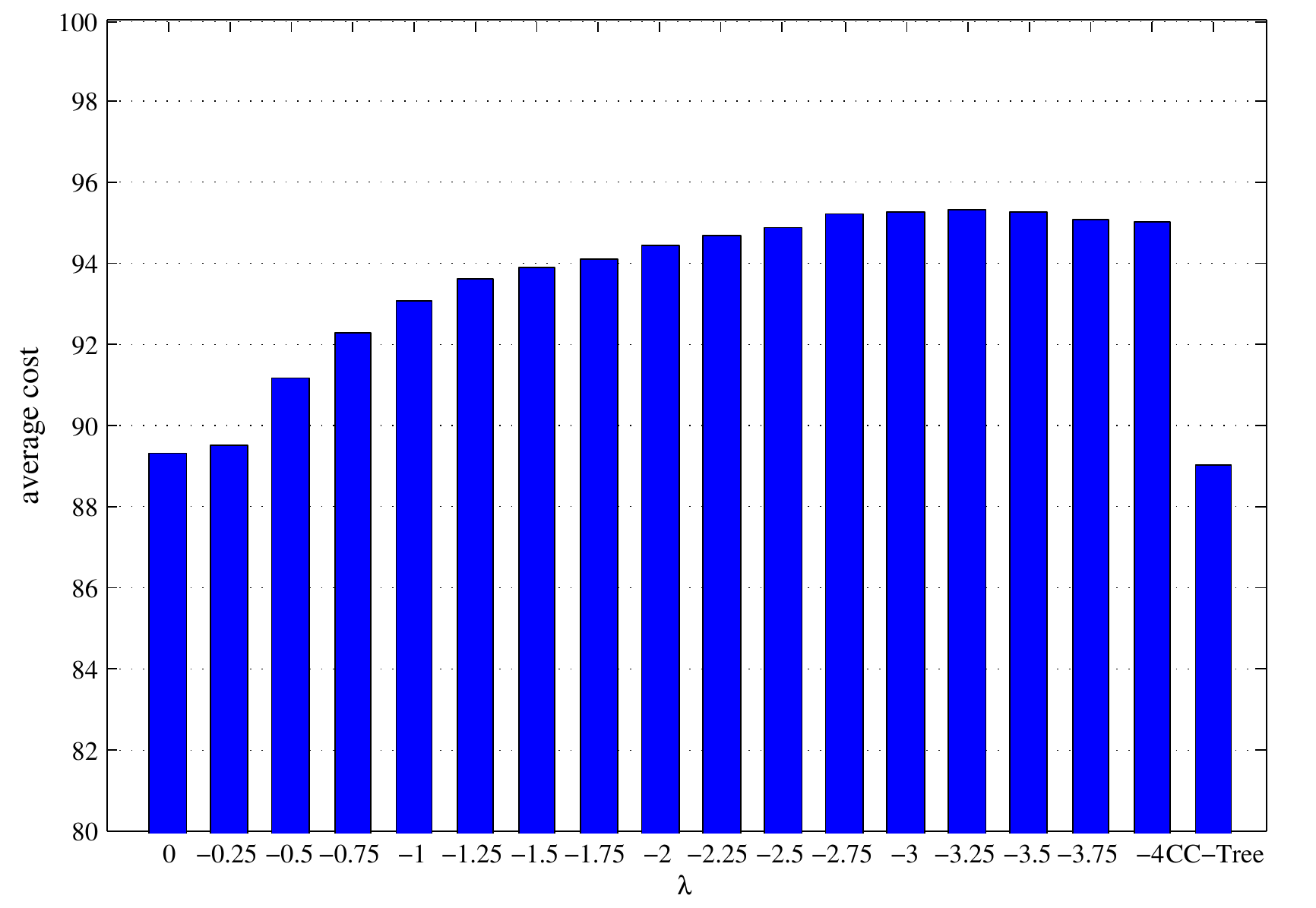}}\\
\subfloat[]{
    \label{fig:ParaComCostN}
    \includegraphics[scale=0.45]{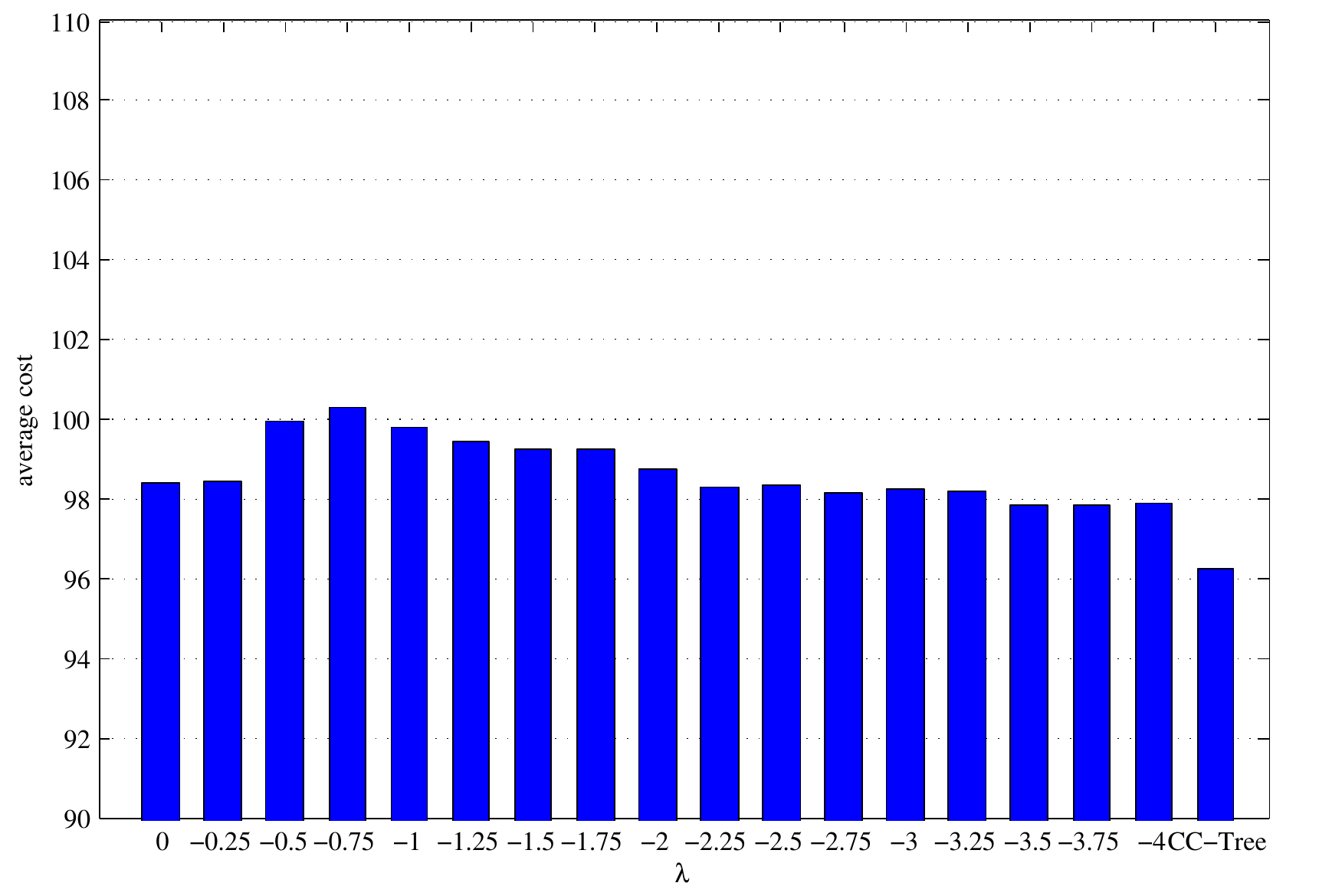}}\\
\subfloat[]{
    \label{fig:ParaComCostP}
    \includegraphics[scale=0.45]{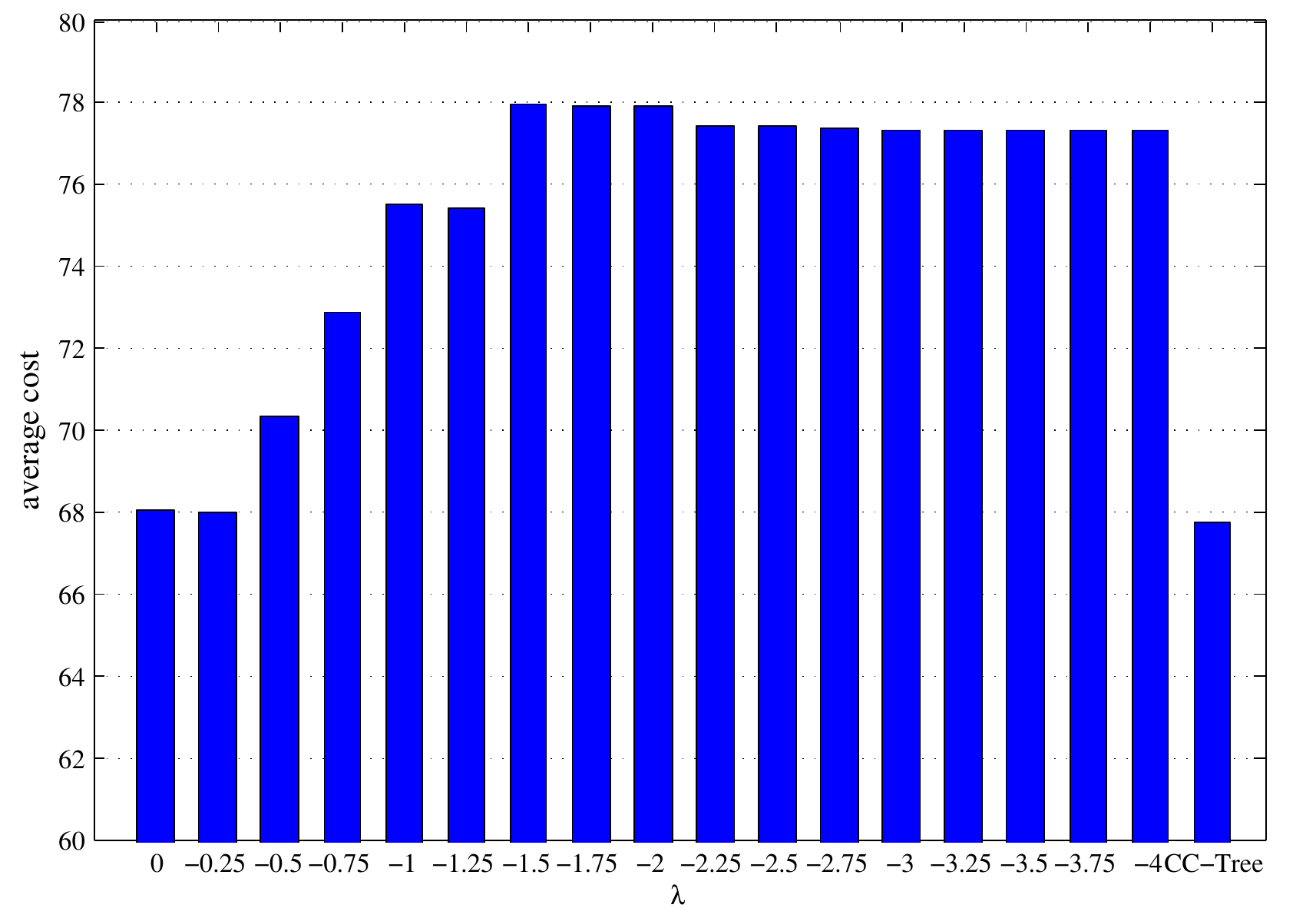}}\\
\caption{Parameter comparison on \texttt{diabetes} dataset with three distributions: (a) Uniform (b) Normal (c) Pareto}
\label{fig:ParaComparCost}
\end{figure}

This section of experiment setup is to study the influence of the $\lambda$ value and the competition strategy.
Our algorithm is tested on dataset \texttt{diabetes} 1000 times.
$\lambda$ value is set to be a fraction between -4 and 0 with the step length 0.25.
We list the results in Figure \ref{fig:ParaComparWintime} about win times and Figure \ref{fig:ParaComparCost} about average cost.
In Figure \ref{fig:ParaComN}, when the $\lambda$ value is 0, the win times is 0.

\subsubsection{Pruning comparison}

\begin{figure}[tb]
\centering
\subfloat[]{
    \label{fig:ParaComU}
    \includegraphics[scale=0.5]{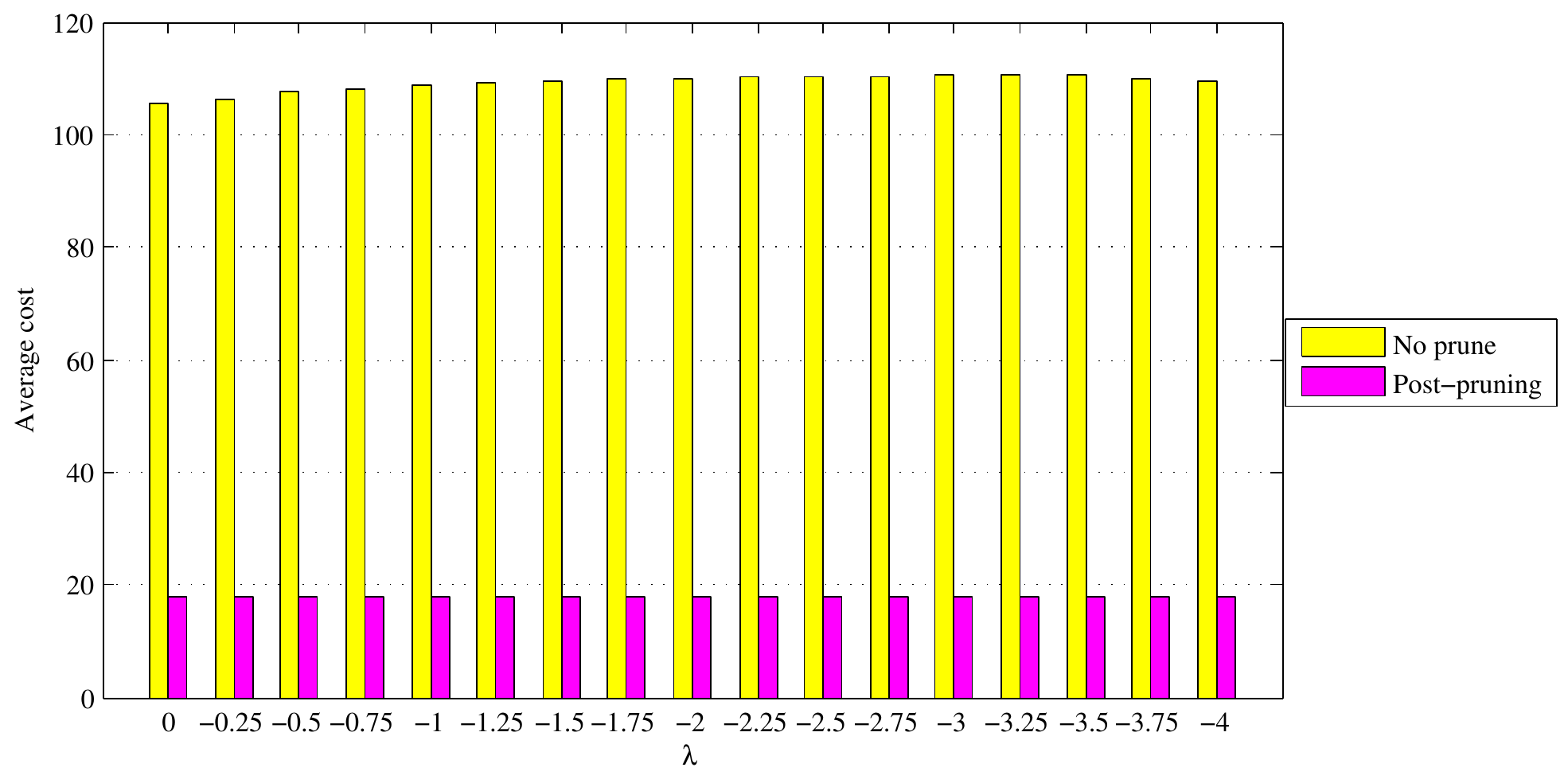}}\\
\subfloat[]{
    \label{fig:ParaComN}
    \includegraphics[scale=0.5]{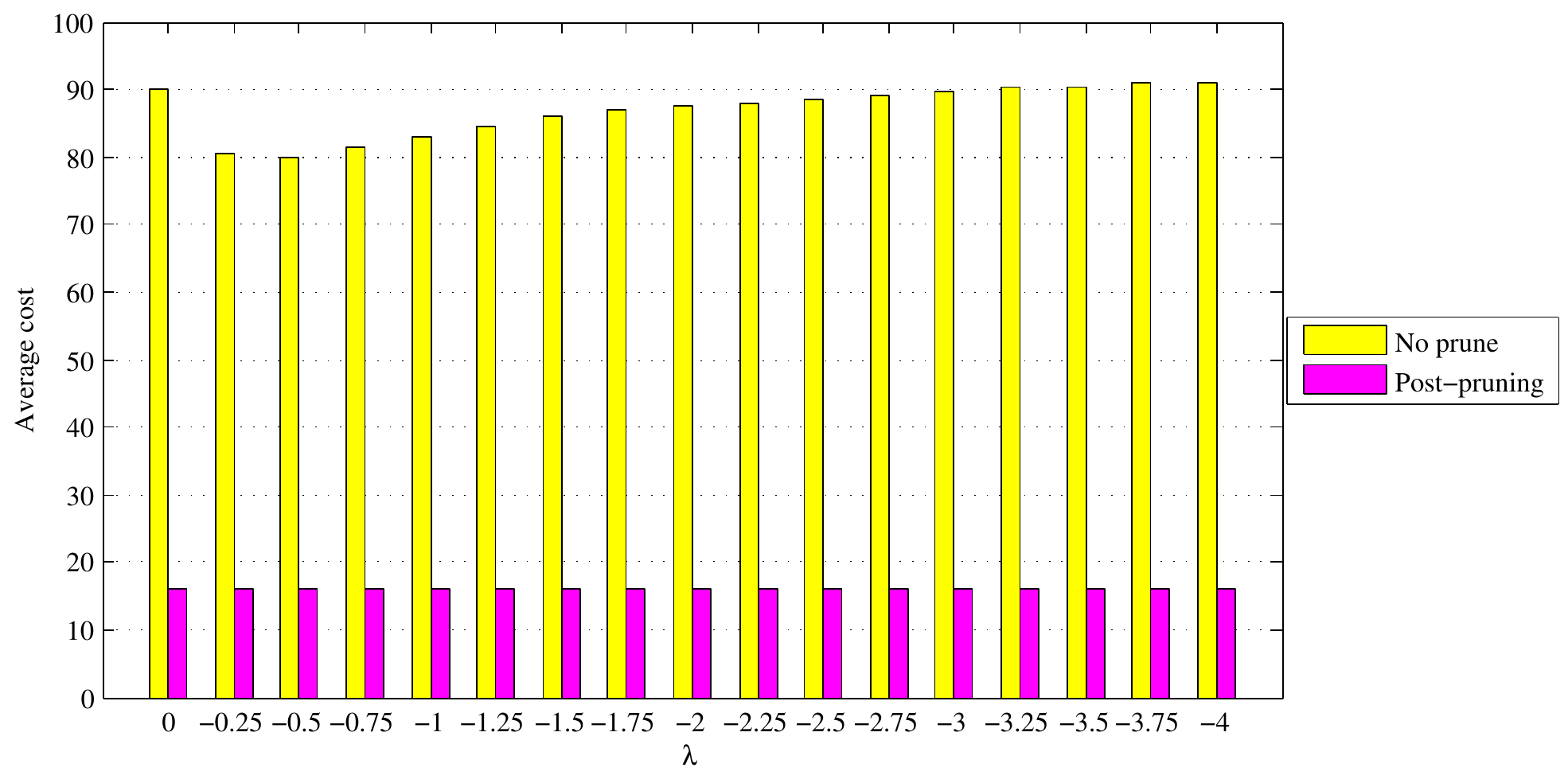}}\\
\subfloat[]{
    \label{fig:ParaComP}
    \includegraphics[scale=0.5]{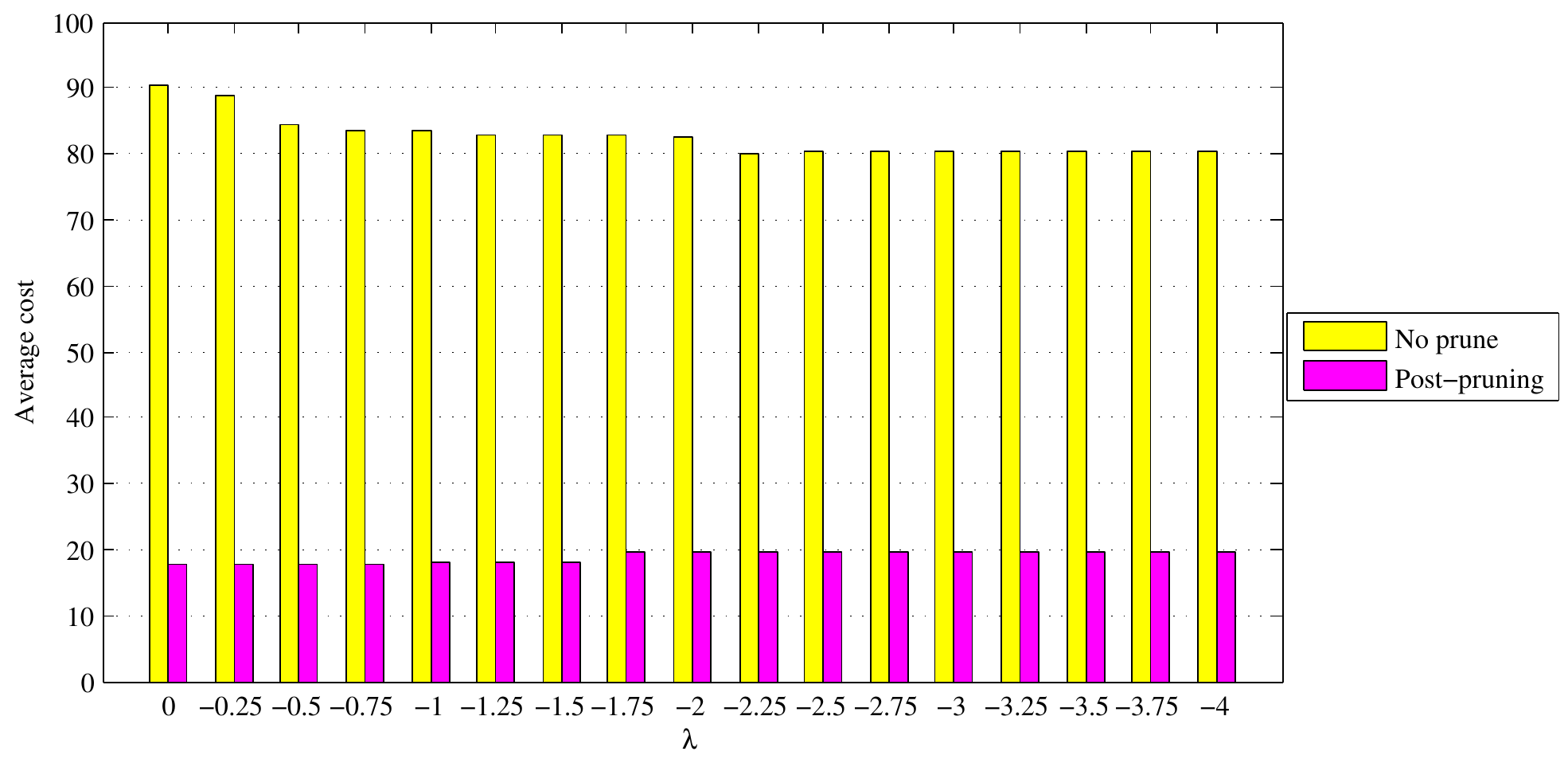}}\\
\caption{Pruning comparison on \texttt{diabetes} dataset with three distributions: (a) Uniform (b) Normal (c) Pareto}
\label{fig:PruneComparCost}
\end{figure}

\begin{figure*}[tb]
    \begin{center}
    \includegraphics[width=4.5in]{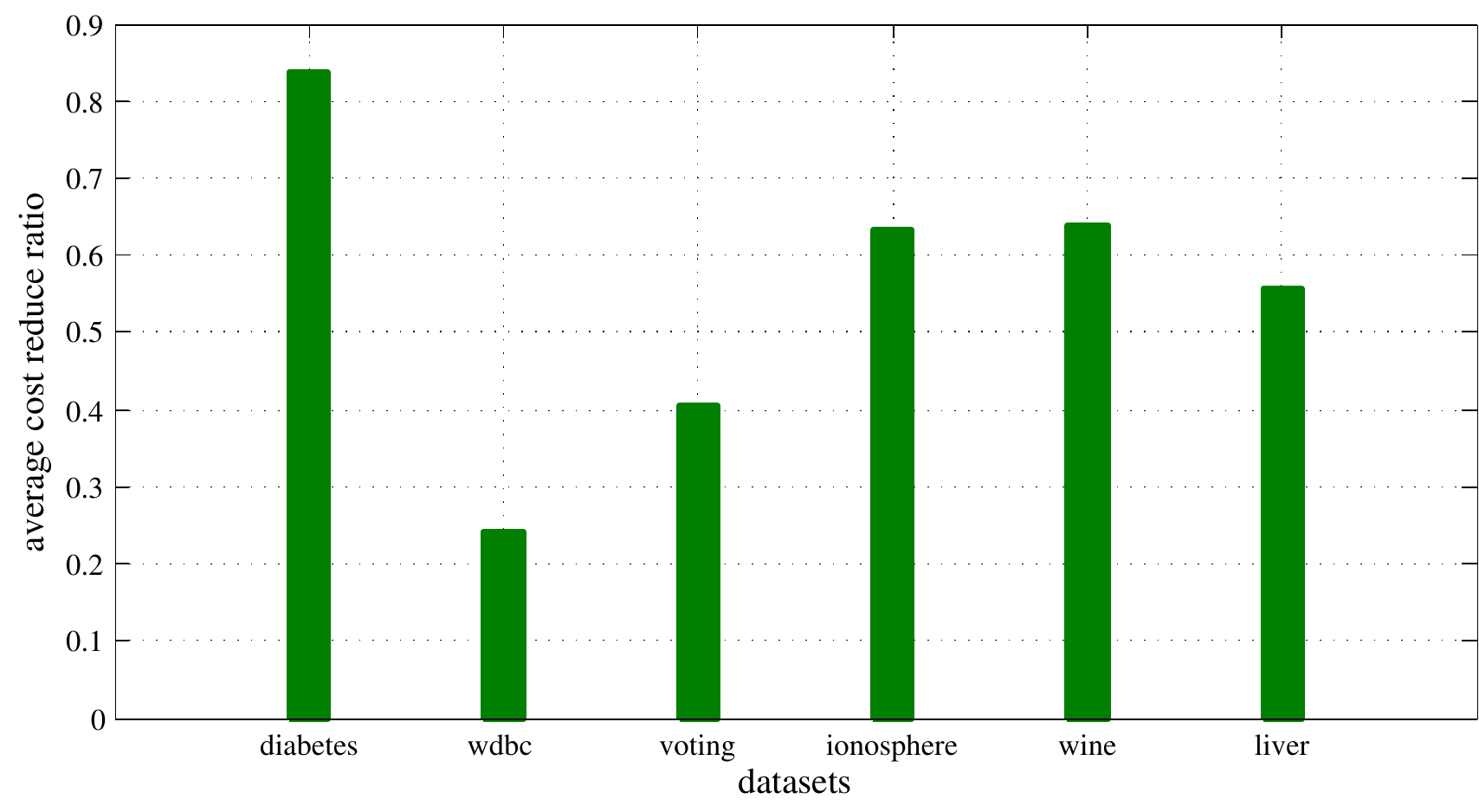}
    \caption{Average reduction ratio of cost on different datasets}
    \label{figure: ParaComparisonCost}
    \end{center}
\end{figure*}

For evaluating the effect of post-pruning technique, we run our algorithm on the same dataset with no pruning and post-pruning.
The specific experimental results on dataset \texttt{diabetes} are shown in Figure \ref{fig:PruneComparCost}.
The results on different datasets with Uniform distribution of the test cost are shown as Figure \ref{figure: ParaComparisonCost}.

\subsubsection{Comparison between training set and testing set}

For studying the stability of our algorithm, we run and evaluate it on training dataset and testing dataset.
The metric is the average value of average cost of decision trees.

\begin{figure}[tb]
\centering
\subfloat[No prune]{
    \label{subfigure:TrainsetCost}
    \includegraphics[scale=0.5]{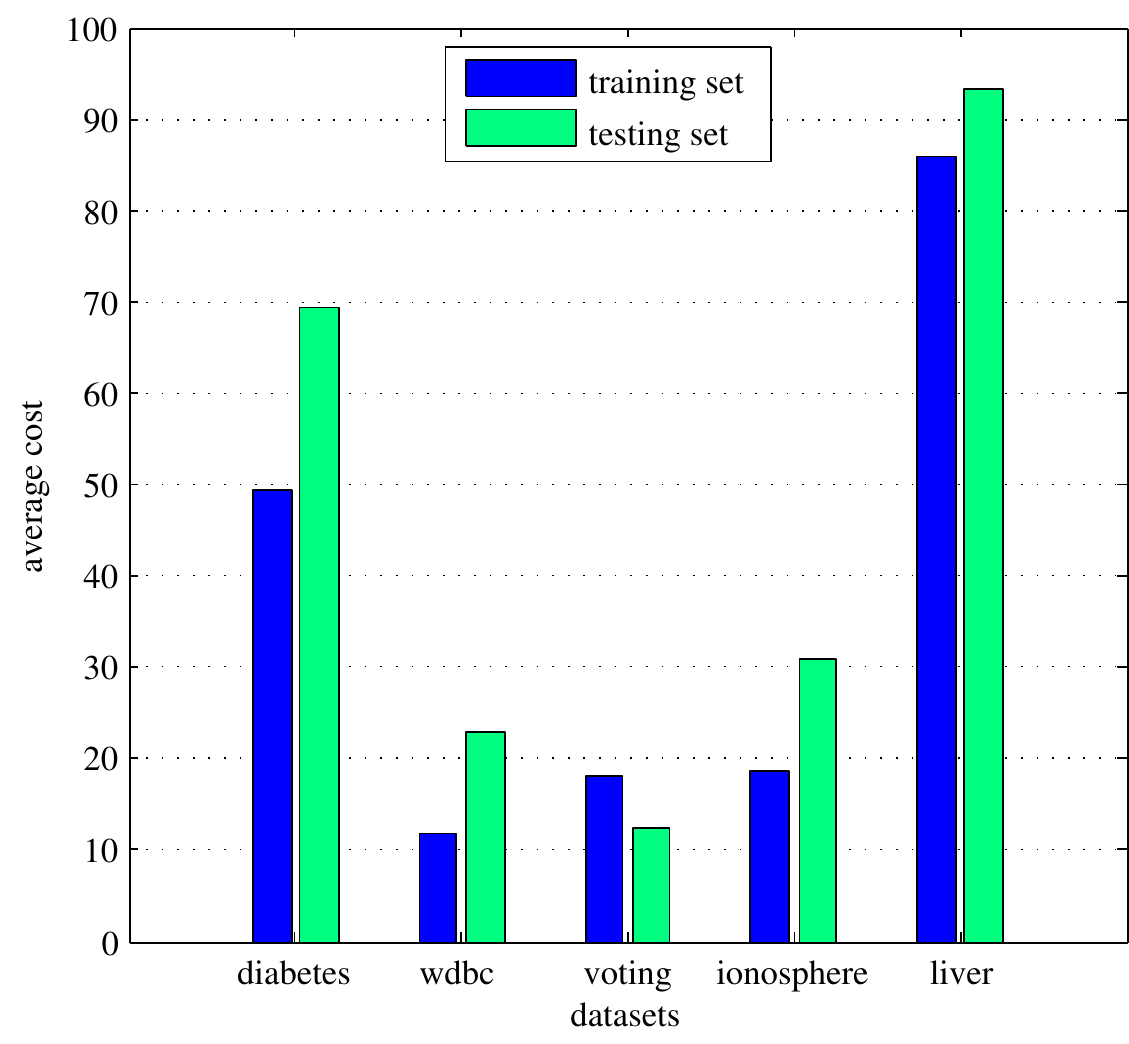}}
\subfloat[Post-pruning]{
    \label{subfigure:TestsetCost}
    \includegraphics[scale=0.52]{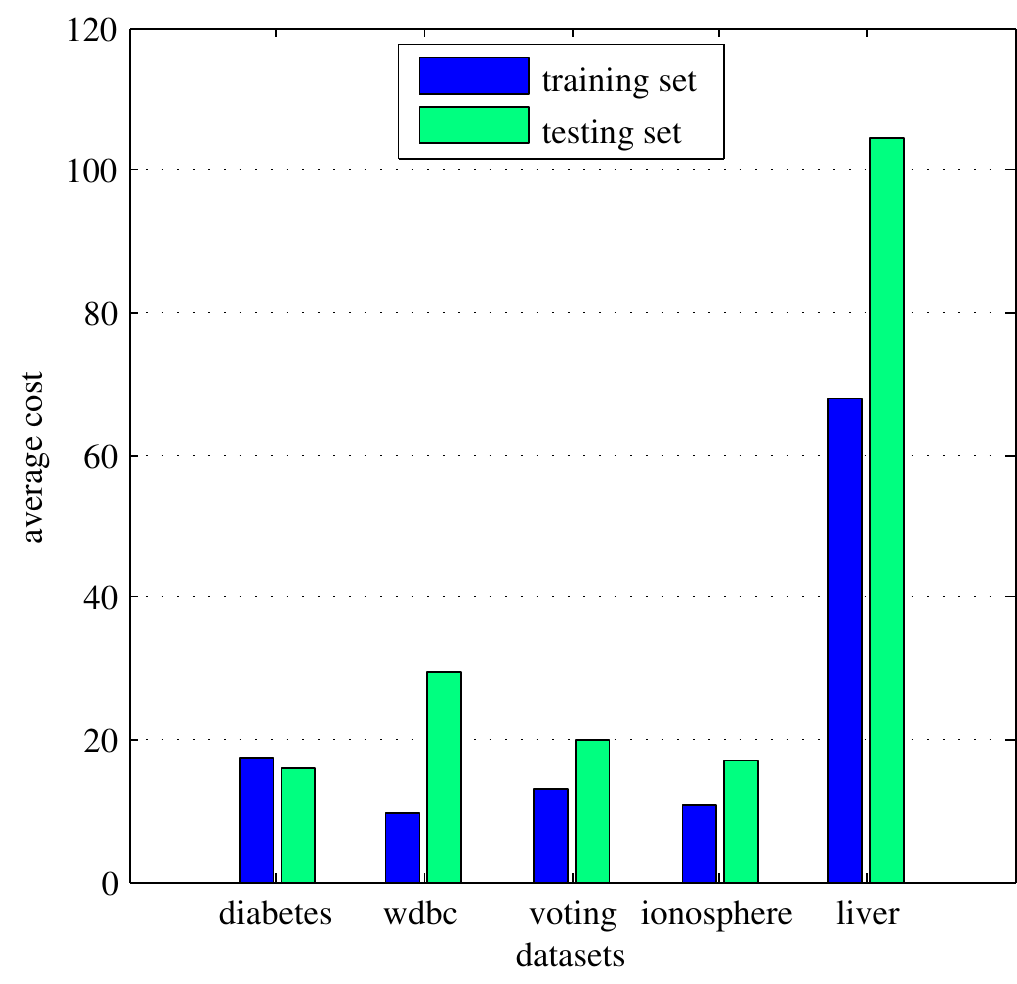}}\\
\caption{Comparison between training set and testing set}
\label{fig:TrainTestComparison}
\end{figure}

Experimental results are shown in Figure \ref{fig:TrainTestComparison}.
From this figure, we can see that the difference of effect between training set and testing set is not significantly.
We can conclude that our algorithm is stable with no overfitting.
All figures in this paper are produced by Matlab \cite{Chen2007Matlab}.

\subsection{Result analysis}

From the experimental results, we can answer the questions proposed in the first of this section.

(1) Experimental results illustrate that our algorithm is stable without overfitting.
It means that our algorithm can obtain similar effect on both training dataset and testing dataset.
We can conclude this statement by Figure \ref{fig:TrainTestComparison}.

(2) Experimental results show that no one $\lambda$ value setting is always optimal on different test cost settings.
It is unreasonable that we use a fixed $\lambda$ value setting.

(3) The competition strategy improves the effectiveness of our algorithm significantly.
The best trees among all the $\lambda$ value settings on the training dataset are best on the testing dataset frequently, but not always.
On the dataset \texttt{diabetes} with three different distributions, it obtains the best result in about 70\%, 80\% and 90\% of the time respectively.
The comparison experiment results on dataset \texttt{diabetes} are shown as Figures \ref{fig:ParaComparWintime} and \ref{fig:ParaComparCost}.
On other datasets, results are similar.

(4) Post-pruning technique reduces the average cost effectively.
The average deceasing rates of cost are about 84\%, 81\% and 55\% with three distributions.
The strategy also decreases the difference among all $\lambda$ value settings.
The comparison experimental results between no prune and post-pruning are shows as Figure \ref{fig:PruneComparCost}.
When the distribution is Unform and Normal, the average cost of the decision trees after post-pruning are the same among all $\lambda$ settings.
The reason is that decision trees with the same number of nodes are generated.
Figure \ref{equation:AverageReduceRatio} reveals the average cost reduction involving the effect of pruning strategy.
Results on 6 different datasets illustrate that the post-pruning technique is effective.


  %
  %
\section{Conclusions}\label{section: conclusion}

In this paper, we have designed a decision tree algorithm based on C4.5 named CC-C4.5.
We adopt the algorithm to deal with cost sensitive tree issue on numeric dataset.
We design a heuristic function based on the test cost and the information gain ratio.
Post-pruning technique is proposed to reduce the average cost of obtained decision trees.
Our algorithm also uses the competition strategy to select the best tree on training dataset to classify.
Experimental results indicate that our algorithm performs stable on training dataset and testing dataset.
Post-pruning is an effective technique to decrease the average cost of the decision trees.
In many cases, the competition strategy can obtain a decision tree with little cost.

  %
  %
\section*{Acknowledgement}
This work is supported in part by the National Natural Science Foundation of China under Grant No. 61170128,
the Natural Science Foundation of Fujian Province, China, under Grant Nos. 2011J01374 and 2012J01294.

  %
  %
\section*{Reference}
\bibliographystyle{elsarticle-num}

%

\end{document}